# Learning Deep Generative Spatial Models for Mobile Robots

Andrzej Pronobis, Rajesh P. N. Rao

*Abstract*— We propose a new probabilistic framework that allows mobile robots to autonomously learn deep, generative models of their environments that span multiple levels of abstraction. Unlike traditional approaches that combine engineered models for low-level features, geometry, and semantics, our approach leverages recent advances in Sum-Product Networks (SPNs) and deep learning to learn a single, universal model of the robot's spatial environment. Our model is fully probabilistic and generative, and represents a joint distribution over spatial information ranging from low-level geometry to semantic interpretations. Once learned, it is capable of solving a wide range of tasks: from semantic classification of places, uncertainty estimation, and novelty detection, to generation of place appearances based on semantic information and prediction of missing data in partial observations. Experiments on laser-range data from a mobile robot show that the proposed universal model obtains performance superior to state-of-the-art models fine-tuned to one specific task, such as Generative Adversarial Networks (GANs) or SVMs.

## I. INTRODUCTION

The ability to acquire and represent spatial knowledge is fundamental for mobile robots operating in large, unstructured environments. Such knowledge exists at multiple levels of abstraction, from robot's sensory data, through geometry and appearance, up to high-level semantic descriptions. Experiments have demonstrated that robots can leverage knowledge at all levels to better perform in the real-world [1].

Traditionally, robotic systems utilize an assembly of independent spatial models [2], which exchange information in a limited fashion. This includes engineered feature extractors and combinations of machine learning techniques, making integration with planning and decision making difficult. At the same time, the recent success of deep learning proves that replacing multiple representations with a single integrated model can lead to a drastic increase in performance [3][4]. As a result, deep models have also been applied to spatial modeling tasks, such as place classification and semantic mapping [5][6]. Yet, the problem was always framed as one of classification, where sensory data is fed to a convolutional neural network (CNN) to obtain semantic labels.

In contrast, in this work our goal is not only to unify multiple representations into a single model, but also to demonstrate that the role of a spatial model can go beyond classification. To this end, we propose the Deep Generative Spatial Model (DGSM), a probabilistic model which learns a joint distribution between a low-level representation of the geometry of local environments (places) and their semantic interpretation. Our model leverages Sum-Product Networks (SPNs), a novel probabilistic deep architecture [7][8].

SPNs have been shown to provide state-of-the-art results in several domains [9][10][11]. However, their potential has not previously been exploited in robotics. DGSM consists of an SPN with a unique structure designed to hierarchically represent the geometry and semantics of a place from the perspective of a mobile robot acting in its environment. To this end, the network represents place geometry using a robot-centric, polar grid, where the nearby objects are captured in more detail than more distant context. On top of the place geometry, we propose a unique network structure which combines domain knowledge with random network generation (which can be seen as a form of structure learning) for parts of the network modeling complex dependencies.

DGSM is generative, probabilistic, and therefore universal. Once learned, it enables a wide range of inferences. First, it can be used to infer a semantic category of a place from sensory input together with probability representing uncertainty. The probabilistic output provides rich information to a potential planning or decision-making subsystem. However, as shown in this work, it can also be used to detect novel place categories. Furthermore, the model reasons jointly about the geometry of the world and its semantics. We exploit that property for two tasks: to generate prototypical appearances of places based on semantic information and to infer missing geometry information in partial observations. We use laser-range data to capture the geometry of places; however the performance of SPNs for vision-based tasks [7][9] indicates that the model should also accommodate 3D and visual information without changing the general architecture.

Our goal is to demonstrate the potential of DGSM, and deep generative models in general, to spatial modeling in robotics. Therefore, we present results of four different experiments addressing each of the inference tasks. In each experiment, we compare our universal model to an alternative approach that is designed for and fine-tuned to a specific task. First, for semantic categorization, we compare to a well-established Support Vector Machine (SVM) model learned on widely used geometrical laser-range features [12][13]. Second, we benchmark novelty detection against one-class SVM trained on the same features. In both cases, DGSM offers superior accuracy. Finally, we compare the generative properties of our model to Generative Adversarial Networks (GANs) [14][15] on the two remaining inference tasks, reaching state-of-the-art accuracy and superior efficiency beyond real-time. This serves as a benchmark but also demonstrates the use of GANs for spatial modeling in robotics. Importantly, to open doors for the use of SPNs in a broader range of applications, we introduce *LibSPN* [16], a new general library for SPN learning and inference.

The authors are with Paul G. Allen School of Computer Science & Engineering, University of Washington, Seattle, WA, USA. A. Pronobis is also with Robotics, Perception and Learning Lab, KTH, Stockholm, Sweden. {pronobis,rao}@cs.washington.edu. This work was supported by Office of Naval Research (ONR) grant no. N00014-13-1-0817 and Swedish Research Council (VR) project 2012-4907 SKAEENet. The help by Kaiyu Zheng and Kousuke Ariga is gratefully acknowledged.

## II. RELATED WORK

Representing semantic spatial knowledge is a broadly researched topic, with many solutions employing vision [17][18][2][5]. Images clearly carry rich information about semantics; however, they are also affected by changing environment conditions. At the same time, robotics researchers have seen advantages of using range data that are much more robust in real-world settings and easier to process in real time. In this work, we focus on laser-range data, as a way of introducing and evaluating a new spatial model as well as a recently proposed deep architecture.

Laser-range data have been extensively used for place classification and semantic mapping, and many traditional, handcrafted representations have been proposed. Buschka et al. [19] contributed a simple method that incrementally divided grid maps of indoor environments into two classes of open spaces (rooms and corridors). Mozos et al. [12] applied AdaBoost to create a classifier based on a set of manually designed geometrical features to classify places into rooms, corridors and doorways. In [20], omnidirectional vision was combined with laser data to build descriptors, called fingerprints of places. Finally, in [13], SVMs have been applied to the geometrical features of Mozos et al. [12] leading to significant improvement in performance over the original AdaBoost. That approach has been further integrated with visual and object cues for semantic mapping in [2].

Deep learning and unsupervised feature learning, after many successes in speech recognition and computer vision [3], entered the field of robotics with superior performance in object recognition [21][22] and robot grasping [23][4]. The latest work in place classification also employs deep approaches. In [5], deep convolutional network (CNN) complemented with a series of one-vs-all classifiers is used for visual semantic mapping. In [6], CNNs are used to classify grid maps built from laser data into 3 classes: room, corridor, and doorway. In these works, deep models are used exclusively for classification, and use of generative models has not been explored. In contrast, we propose a universal probabilistic generative model, and demonstrate its usefulness for multiple robotics tasks, including classification.

Several generative, deep architectures have recently been proposed, notably Variational Autoencoders [24], Generative Adversarial Networks [14], and Sum-Product Networks [8][7][9]. GANs have been shown to produce high-quality generative representations of visual data [15], and have been successfully applied to image completion [25]. SPNs, a probabilistic model, achieved promising results for varied applications such as speech [10] and language modeling [26], human activity recognition [11], and image classification [9] and completion [7], but have not been used in robotics. In this work, we exploit the universality and efficiency of SPNs to propose a single spatial model able to solve a wide range of inference problems relevant to a mobile robot. Furthermore, inspired by their results in other domains, we also evaluate GANs (when applicable). This serves as a comparison and a demonstration of GANs on a new application.

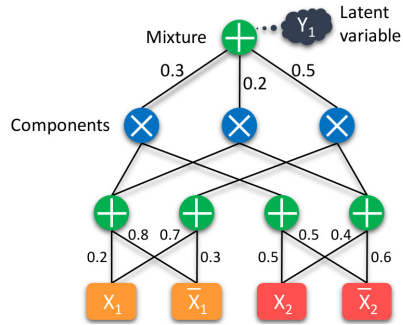

Fig. 1: An SPN for a naive Bayes mixture model $P(X_1, X_2)$, with three components over two binary variables. The bottom layer consists of indicators for $X_1$ and $X_2$. Weighted sum nodes, with weights attached to inputs, are marked with $+$, while product nodes are marked with $\times$. $Y_1$ represents a latent variable marginalized out by the root sum.

## III. SUM-PRODUCT NETWORKS

Sum-product networks are a recently proposed probabilistic deep architecture with several appealing properties and solid theoretical foundations [8][7][9]. One of the primary limitations of probabilistic graphical models is the complexity of their partition function, often requiring complex approximate inference in the presence of non-convex likelihood functions. In contrast, SPNs represent joint or conditional probability distributions with partition functions that are guaranteed to be tractable and involve a polynomial number of sum and product operations, permitting exact inference. SPNs are a deep, hierarchical representation, capable of representing context-specific independence and performing fast, tractable inference on high-treewidth models. While not all probability distributions can be encoded by polynomial-sized SPNs, recent experiments in several domains show that the class of distributions modeled by SPNs is sufficient for many real-world problems, offering real-time efficiency.

As shown in Fig. 1, on a simple example of a naive Bayes mixture model, an SPN is a generalized directed acyclic graph composed of weighted sum and product nodes. The sums can be seen as mixture models over subsets of variables, with weights representing mixture priors. Products can be viewed as features or mixture components. The latent variables of the mixtures can be made explicit and their values inferred. This is often done for classification models, where the root sum is a mixture of sub-SPNs representing classes. The bottom layers effectively define features reacting to certain values of indicators[1] for the input variables.

Formally, following Poon & Domingos [7], we can define an SPN as follows:

*Definition 1:* An SPN over variables $X_1, \ldots, X_V$ is a rooted directed acyclic graph whose leaves are the indicators $(X_1^1, \ldots, X_1^I), \ldots, (X_V^1, \ldots, X_V^I)$ and whose internal nodes are sums and products. Each edge $(i, j)$ emanating from a sum node $i$ has a non-negative weight $w_{ij}$. The value of a product node is the product of the values of its children. The value of a sum node is $\sum_{j \in Ch(i)} w_{ij} v_j$, where $Ch(i)$ are the children of $i$ and $v_j$ is the value of node $j$. The value of an SPN $S[X_1, \ldots, X_V]$ is the value of its root.

---
[1]Indicator is a binary variable set to 1 when the corresponding categorical variable takes a specific value. For using continuous input variables, see [7].

Not all architectures consisting of sums and products result in a valid probability distribution. However, following simple constraints on the structure of an SPN will guarantee validity (see [7], [8]). When the weights of each sum node are normalized to sum to 1, the value of a valid SPN $S[X_1^1, \ldots, X_V^I]$ is equal to the normalized probability $P(X_1, \ldots, X_V)$ of the distribution modeled by the network [8].

### A. Generating SPN structure

The structure of the SPN determines the group of distributions that can be learned. Therefore, most previous works [9][11][26] relied on domain knowledge to design the appropriate structure. Furthermore, several structure learning algorithms were proposed [27][28] to discover independencies between the random variables in the dataset, and structure the SPN accordingly. In this work, we experiment with a different approach, originally hinted at in [7], which generates a random structure, as in random forests. Such an approach has not been previously evaluated. Our experiments demonstrate that it can lead to very good performance and can accommodate a wide range of distributions. Additionally, after parameter learning, the generated structure can be pruned by removing edges associated with weights close to zero. This can be seen as a form of structure learning.

To obtain the random structure, we recursively generate nodes based on multiple random decompositions of a set of random variables into multiple subsets until each subset is a singleton. As illustrated in Fig. 2 (middle, Level 1), at each level the current set of variables to be decomposed is modeled by multiple mixtures (green nodes), and each subset of the decomposition is also modeled by multiple mixtures (green nodes one level below). Product nodes (blue) are used as an intermediate layer and act as features detecting particular combinations of mixtures representing each subset. The top mixtures of each level mix outputs of all product nodes at that level. The same set of variables can be decomposed into subsets in multiple random ways (e.g. there are two decompositions at the top of Fig. 2).

### B. Inference and Learning

Inference in SPNs is accomplished by a single pass through the network. Once the indicators are set to represent the evidence, the upward pass will yield the probability of the evidence as the value of the root node. Partial evidence (or missing data) can easily be expressed by setting all indicators for a variable to 1. Moreover, since SPNs compute a network polynomial [29], derivatives computed over the network can be used to perform inference for modified evidence without recomputing the whole SPN. Finally, it can be shown [8] that MPE inference in a certain class of SPNs (selective) can be performed by replacing all sum nodes with max nodes while retaining the weights. Then, the indicators of the variables for which the MPE state is inferred are all set to 1 and a standard upward pass is performed. A downward pass then follows, which recursively selects the highest valued child of each sum (max) node and all children of a product node. The indicators selected by this process indicate the MPE state of the variables. In general SPNs, this algorithm yields an approximation of the MPE state.

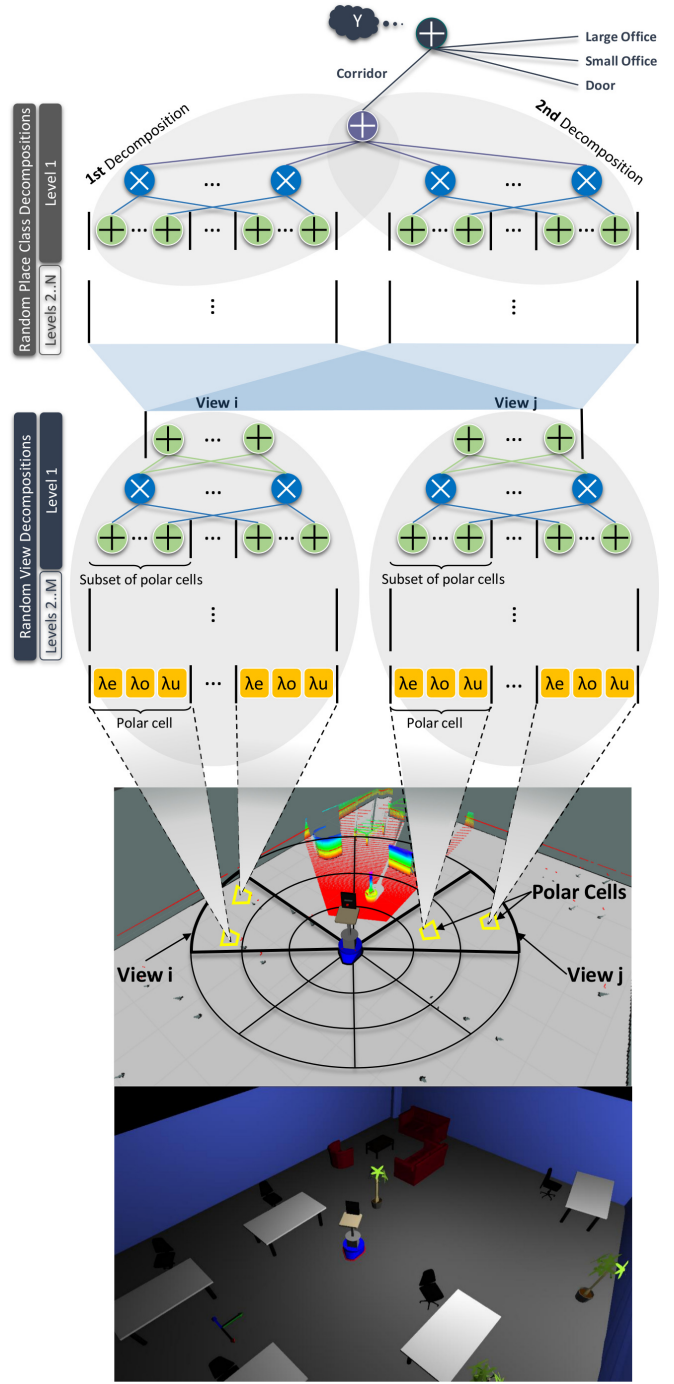

Fig. 2: The structure of the SPN implementing our spatial model. The bottom images illustrate a robot in an environment and a robot-centric polar grid formed around the robot. The SPN is built on top of the variables representing the occupancy in the polar grid.

SPNs lend themselves to be learned generatively [7] or discriminatively [9] using Expectation Maximization (EM) or gradient descent. In this work, we employ hard EM to learn the weights, which was shown to work well for generative learning [7]. As is often the case for deep models, the gradient quickly diminishes as more layers are added. Hard EM overcomes this problem, permitting learning of SPNs with hundreds of layers. Each iteration of the EM learning consists of an MPE inference of the implicit latent variables of each sum with training samples set as evidence (E step), and an update of weights based on the inference results (M step, for details, see [7]). We achieved best results by modifying the MPE inference to use sums instead of maxes during the upwards pass, while selecting the max valued child during the downward pass. Furthermore, we performed additive smoothing when updating the weights corresponding to a Dirichlet prior and terminated learning after 300 iterations. No additional learning parameters are required.

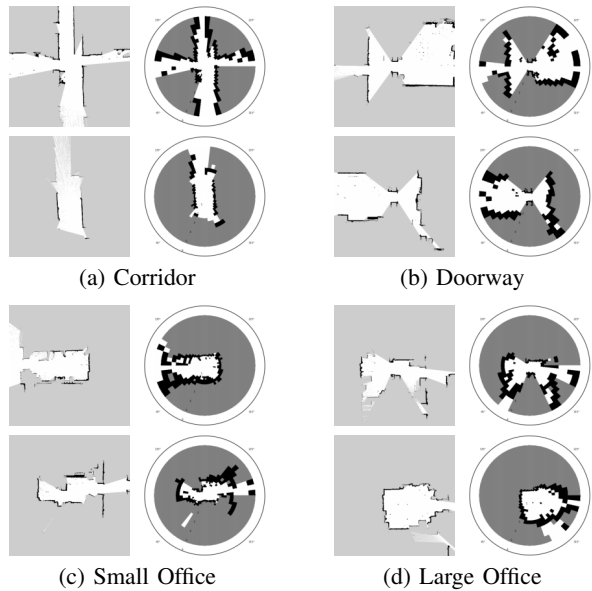

Fig. 3: Local environment observations used in our experiments, expressed as Cartesian and polar occupancy grids, for examples of places of different semantic categories.

## IV. DEEP GENERATIVE SPATIAL MODEL (DGSM)

### A. Representing Local Environments

DGSM is designed to support real-time, online spatial reasoning on a mobile robot. A real robot almost always has access to a stream of observations of the environment. Thus, as the first step, we perform spatio-temporal integration of the sensory input. We rely on laser-range data, and use a particle-filter grid mapping [30] to maintain a robot-centric map of 5m radius around the robot. During acquisition of the dataset used in our experiments, the robot was navigating a fixed path through a new environment, while continuously integrating data gathered using a single laser scanner with 180° FOV. This still results in partial observations of the surroundings (especially when the robot enters a new room), but helps to assemble a more complete representation over time.

Our goal is to model the geometry and semantics of a local environment only. We assume that larger-scale spatial model will be built by integrating multiple models of local places. Thus, we constrain the observation of a place to the information visible from the robot (structures that can be raytraced from the robot's location). As a result, walls occlude the view and the local map mostly contains information from a single room. In practice, additional noise is almost always present, but is averaged out during learning of the model. Examples of such local environment observations can be seen in Fig. 3.

Next, each local observation is transformed into a robot-centric polar occupancy grid (compare polar and Cartesian grids in Fig. 3). The resulting observation contains higher-resolution details closer to the robot and lower-resolution context further away. This focuses the attention of the model on the nearby objects. Higher resolution of information closer to the robot is important for understanding the semantics of the robot's exact location (for instance when the robot is at a doorway). However, it also relates to how spatial information is used by a mobile robot when planning and executing actions. It is in the vicinity of the robot that higher accuracy of spatial information is required. A similar principle is exploited by many navigation components, which use different resolution of information for local and global path planning. Additionally, such a representation corresponds to the way the robot perceives the world because of the limited resolution of its sensors. Our goal is to use a similar strategy when representing 3D and visual information, by extending the polar representation to 3 dimensions. Finally, a high-resolution map of the complete environment can be largely recovered by integrating stored or inferred polar observations over the path of the robot. We built polar grids of radius of 5m, with an angle step of 6.4 degrees and grid resolution decreasing with the distance from the robot.

### B. Architecture of DGSM

The architecture of DGSM is based on a generative SPN illustrated in Fig. 2. The model learns a probability distribution $P(Y, X_1, \ldots, X_C)$, where $Y$ represents the semantic category of a place, and $X_1, \ldots, X_C$ are input variables representing the occupancy in each cell of the polar grid. Each occupancy cell is represented by three indicators in the SPN (for empty, occupied and unknown space). These indicators constitute the bottom of the network (orange nodes).

The structure of the model is partially designed based on domain knowledge and partially generated according to the algorithm described in Sec. III-A. The resulting model is a single SPN assembled from three levels of sub-SPNs. We begin by splitting the polar grid equally into 8 45-degree *views*. For each view, we generate a random sub-SPN by recursively building a hierarchy of decompositions of subsets of polar cells in the view. Then, on top of all the sub-SPNs representing the views, we generate an SPN representing complete place geometries for each place class. Finally, the sub-SPNs for place classes are combined by a sum node forming the root of the network. The latent variable associated with that sum node is made explicit as $Y$ and represents the semantic class of a place.

Sub-dividing the representation into views allows us to use networks of different complexity for representing lower-level view features and high-level structure of a place. In our experiments, when representing views, we recursively decomposed the set of polar cells using a single random decomposition, into 2 cell sub-sets, and generated 4 mixtures to model each such subset. This procedure was repeated until each subset contained a single variable representing a single cell. To increase the discriminative power of each view representation, we used 14 sums at the top level of the view sub-SPN. These sums are considered input to a randomly generated SPN structure representing a place class. To ensure that each class can be associated with a rich assortment of place geometries, we increased the complexity of the generated structure and performed 4 random decompositions of the sets of mixtures representing views into 5 subsets. The performance of the model does not vary greatly with the structure parameters as long as the generated structure is sufficiently expressive to support learning of dependencies in the data.

Several straightforward modifications to the architecture can be considered. First, the latent variables in the mixtures modeling each view can be made explicit and considered a view or scene descriptor discovered by the learning algorithm. Second, the weights of the network could be shared across views, potentially simplifying the learning process.

*C. Types of Inference*

As a generative model of a joint distribution between low-level observations and high-level semantic phenomena, DGSM is capable of various types of inferences.

First, the model can simply be used to classify observations into semantic categories, which corresponds to MPE inference of $y$: $y^* = \mathrm{argmax}_y P(y|x_1,\ldots,x_C)$. Second, the likelihood of an observation can be used as a measure of novelty and thresholded: $\sum_y P(y,x_1,\ldots,x_C) > t$. We use this approach to separate test observations of classes known during training from observations of unknown classes.

If instead, we condition on the semantic information, we can perform MPE inference over the variables representing occupancy of polar grid cells:

$$x_1^*,\ldots,x_C^* = \underset{x_1,\ldots,x_C}{\mathrm{argmax}}\, P(x_1,\ldots,x_C|y).$$

This leads to generation of prototypical examples for each class. Finally, we can use partial evidence about the occupancy and infer the most likely state of a subset of polar grid cells for which evidence is missing:

$$x_J^*,\ldots,x_C^* = \underset{x_J,\ldots,x_C}{\mathrm{argmax}}\, \sum_y P(y,x_1,\ldots,x_{J-1},x_J,\ldots,x_C)$$

We use this technique to infer missing observations in our experiments.

## V. GANS FOR SPATIAL MODELING

Recently, Generative Adversarial Networks [14] have received significant attention for their ability to learn complex visual phenomena in an unsupervised way [15], [25]. The idea behind GANs is to simultaneously train two deep networks: a generative model $G(z;\theta_g)$ that captures the data distribution and a discriminative model $D(x;\theta_d)$ that discriminates between samples from the training data and samples generated by $G$. The training alternates between updating $\theta_d$ to correctly discriminate between the true and generated data samples and updating $\theta_g$ so that $D$ is fooled. The generator is defined as a function of noise variables $z$, typically distributed uniformly (values from -1 to 1 in our experiments). For every value of $z$, a trained $G$ should produce a sample from the data distribution.

Although GANs have been known to be unstable to train, several architectures have been proposed that result in stable models over a wide range of datasets. Here, we employ one such architecture called DC-GAN [15], which provides excellent results on datasets such as MNIST, LSUN, ImageNet [15] or CelebA [25]. Specifically, we used 3 convolutional layers (of dimensions $18\times18\times64$, $9\times9\times128$, $5\times5\times256$) with stride 2 and one fully-connected layer for $D^2$. We used an analogous architecture based on fractional strided convolutions for $G$. We assumed $z$ to be of size 100. DC-GANs do not use pooling layers, perform batch normalization for both $D$ an $G$, and use ReLU and LeakyReLU activations for $D$ and $G$, respectively. We used ADAM to learn the parameters.

Since DC-GAN is a convolutional model, we could not directly use the polar representation as input. Instead, we used the Cartesian local grid maps directly. The resolution of the Cartesian maps was set to $36\times36$, which is larger then the average resolution of the polar grid, resulting in 1296 occupancy values being fed to the DC-GAN, as compared to 1176 for DGSM. We encoded input occupancy values into a single channel[2] (0, 0.5, 1 for unknown, empty, and occupied).

*A. Predicting Missing Observations*

In [25], a method was proposed for applying GANs to the problem of image completion. The approach first trains a GAN on the training set and then relies on stochastic gradient descent to adjust the value of $z$ according to a loss function $\mathcal{L}(z) = \mathcal{L}_c(z) + \lambda\mathcal{L}_p(z)$, where $\mathcal{L}_c$ is a contextual loss measuring the similarity between the generated and true known input values, while $\mathcal{L}_p$ is a perceptual loss which ensures that the recovered missing values look real to the discriminator. We use this approach to infer missing observations in our experiments. While effective, it requires iterative optimization to infer the missing values. This is in contrast to DGSM, which performs inference using a single up/down pass through the network. We selected the parameter $\lambda$ to obtain the highest ratio of correctly reconstructed pixels.

## VI. EXPERIMENTS

We conducted four experiments corresponding to the inference types described in Sec. IV-C. Importantly, DGSM was trained only once and the same instance of the model was used for all inferences. For each experiment, we compared to a baseline model fine-tuned to the specific sub-problem.

---

[2]We evaluated architectures consisting of 4 conv. layers and layers of different dimensions (depth of the 1st layer ranging from 32 to 256). We also investigated using two and three channels to encode occupancy information. The final architecture results in significantly better completion accuracy.

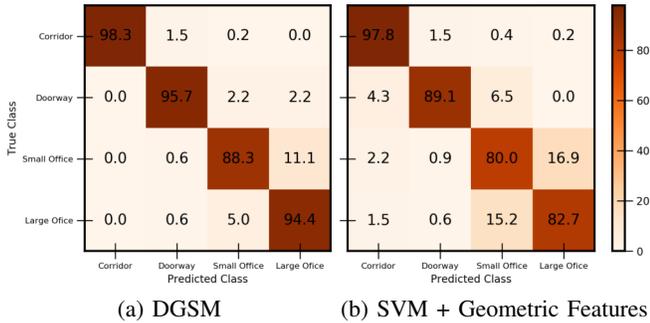

(a) DGSM    (b) SVM + Geometric Features

Fig. 4: Normalized confusion matrices for the task of semantic place categorization.

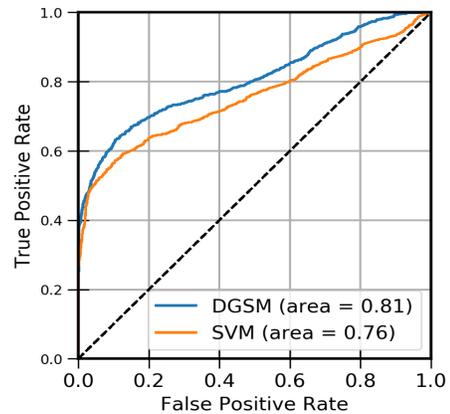

Fig. 5: ROC curves for novelty detection. Inliers are considered positive, while novel samples are negative.

### A. Experimental Setup

Our experiments were performed on laser-range data from the COLD-Stockholm dataset [2]. The dataset contains multiple data sequences captured using a mobile robot navigating with constant speed through four different floors of an office building. On each floor, the robot navigates through rooms of different semantic categories. There are 9 different *large offices*, 8 different *small offices* (distributed across the floors), 4 long *corridors* (1 per floor, with varying appearance in different parts), and multiple examples of places in *doorways*. The dataset features several other room categories: an elevator, a living room, a meeting room, a large meeting room, and a kitchen. However, with only one or two room instances in each category. Therefore, we decided to designate those categories as novel when testing novelty detection and used the remaining four categories for the majority of the experiments. To ensure variability between the training and test sets, we split the data samples four times, each time training the DGSM model on samples from three floors and leaving one floor out for testing. The presented results are averaged over the four splits.

The experiments were conducted using *LibSPN* [16]. SPNs are still a new architecture, and only few, limited domain-specific implementations exist at the time of writing. In contrast, our library offers a general toolbox for structure generation, learning and inference, and enables quick application of SPNs to new domains. It integrates with TensorFlow, which leads to an efficient solution capable of utilizing multiple GPUs, and enables combining SPNs with other deep architectures. The presented experiments are as much an evaluation of DGSM as they are of LibSPN.

### B. Semantic Place Categorization

First, we evaluated DGSM for semantic place categorization and compared it to a well-established model based on an SVM and geometric features [12], [13]. The features were extracted from 360° virtual laser scans raytraced in the original, high-resolution (2cm/pixel) Cartesian grid maps used to form the polar grids for DGSM. To ensure the best SVM result, we used an RBF kernel and selected the kernel and learning parameters directly on the test sets. While early attempts to solve a similar classification problem with deep conv nets exist [6], it is not clear whether they offer performance improvements compared to the SVM-based approach. Additionally, using SVMs allows us to evaluate not only classification, but also novelty detection.

The models were trained on the four room categories and evaluated on observations collected in places belonging to the same category, but on different floors. The normalized confusion matrices are shown in Fig. 4. We can see that DGSM obtains superior results for all classes. The classification rate averaged over all classes (giving equal importance to each class) and data splits is $85.9\% \pm 5.4$ for SVM and $92.7\% \pm 6.2$ for DGSM, with DGSM outperforming SVM for every split. Most of the confusion exists between the small and large office classes. Offices in the dataset often have complex geometry that varies greatly between the rooms.

### C. Novelty Detection

The second experiment evaluated the quality of the uncertainty measure produced by DGSM and its applicability to detecting outliers from room categories not known during training. We used the same DGSM model and relied on the likelihood produced by DGSM to decide if a test sample is from a known or novel category. The cumulative ROC curve over all data splits is shown in Fig. 5.

We compared to a one-class SVM with an RBF kernel trained on the geometric features. The $\nu$ parameter was adjusted to obtain the largest area under the curve (AUC) on the test sets. We can observe that DGSM offers a significantly more reliable novelty signal, with AUC of 0.81 compared to 0.76 for SVM. This result is significant, since to the best of our knowledge, it demonstrates for the first time the usefulness of SPN-based models for novelty detection.

### D. Generating Observations of Places

In this qualitative experiment, our aim was to assess and compare the generative properties of DGSM and GANs by inferring complete, prototypical appearances of places knowing only semantic categories. For DGSM, we conditioned on the semantic class variable and inferred the MPE state of the observation variables. The generated polar occupancy grids are illustrated in Fig. 6a-d. For GANs, we plot samples generated for random values of the noise variables $z$ in Fig. 6e.

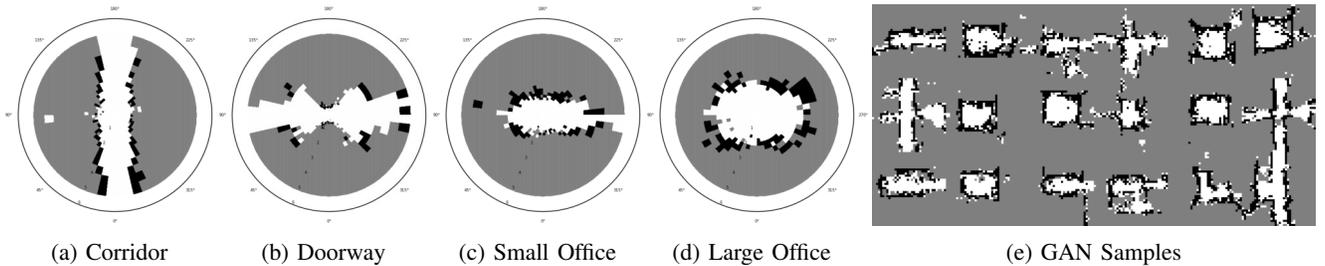

(a) Corridor (b) Doorway (c) Small Office (d) Large Office (e) GAN Samples

Fig. 6: Results of MPE inference over place appearances conditioned on each semantic category for DGSM ((a)-(d)); and place appearance samples generated using GAN (e).

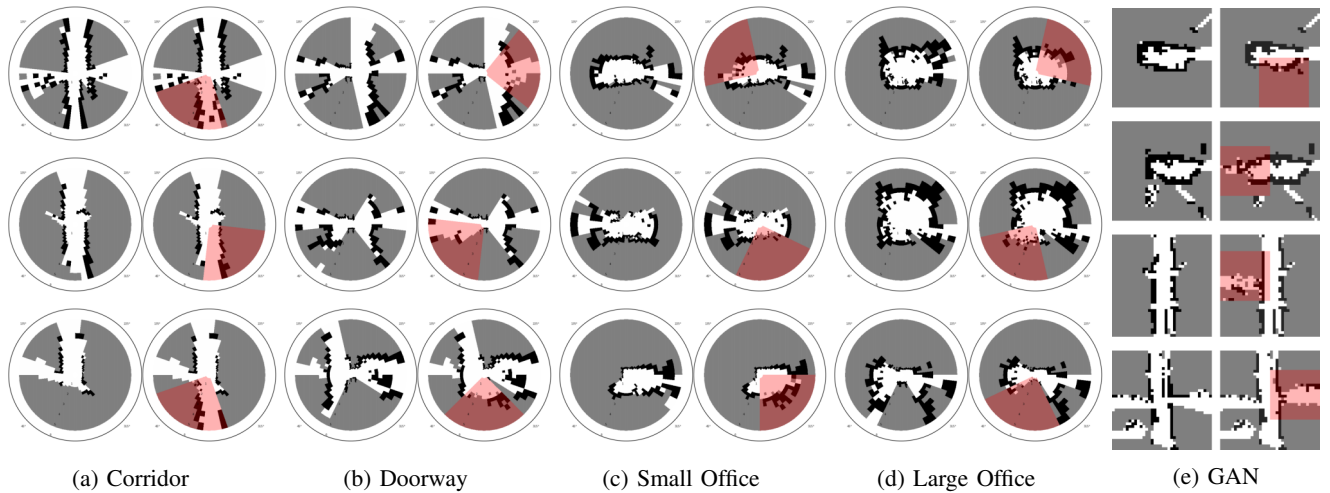

(a) Corridor (b) Doorway (c) Small Office (d) Large Office (e) GAN

Fig. 7: Examples of successful and unsuccessful completions of place observations with missing data: grouped by true semantic category for DGSM ((a)-(d)) and for GAN (e). For each example, a pair of grids is shown, with the true complete grid on the left, and the inferred missing data on the right. The part of the grid that was masked and inferred is highlighted.

We can compare the plots to the true examples depicted in Fig. 3. We can see that each polar grid is very characteristic of the class from which it was generated. The corridor is an elongated structure with walls on either side, and the doorway is depicted as a narrow structure with empty space on both sides. Despite the fact that, as shown in Fig. 3, large variability exists between the instances of offices within the same category, the generated observations of small and large offices clearly have a distinctive size and shape.

While the GAN architecture used for predicting missing observations in unlabeled samples cannot generate samples conditional on semantic category, we still clearly see examples of different room classes and intra-class variations.

### E. Predicting Missing Observations

Our final experiment provides a quantitative evaluation of the generative models on the problem of reconstructing missing values in partial observations of places. To this end, we masked a random chunk of 25% of the grid cells for each test sample in the dataset. In case of DGSM, we masked a random 90-degree view, which corresponds to a rectangular mask in polar coordinates. For GANs, since we use a Cartesian grid, we used a square mask in a random position around the edges of the grid map[3]. For DGSM, all indicators for the masked polar cells were set to 1 to indicate missing evidence and MPE inference followed. For GANs, we used the approach in Sec. V-A.

Fig. 7 shows examples of grids filled with predicted occupancy information to replace the missing values for both models. While the predictions are often consistent with the true values, both models do make mistakes. Analyzing the DGSM results more closely, we see that this typically occurs when the mask removes information distinctive of the place category. Interestingly, in some cases, the unmasked input grid might itself be partial due to missing observations during laser range data acquisition. When the missing observations coincide with a mask, the model will attempt to reconstruct them. Such example can be seen for a polar grid captured in a corridor shown in the bottom left corner of Fig. 7.

Overall, when averaged over all the test samples and data splits, DGSM correctly reconstructs 77.14% ± 1.04 of masked cells, while GANs recover 75.84% ± 1.51. This demonstrates that the models have comparable generative potential, confirming state-of-the-art performance of DGSM.

[3]We considered polar masks on top of Cartesian grid maps. However, this provided a significant advantage to GANs, since most of the masked pixels lay far from the robot, often outside a room, where they are easy to predict.

*F. Discussion and Model Comparison*

The experiments clearly demonstrate the potential of DGSM. Its generative abilities match (and potentially surpass) those of GANs on our problem, while being significantly more computationally efficient. DGSM naturally represents missing evidence and requires only a single upwards and downwards pass through the network to infer missing observations, while GANs required hundreds of iterations, each propagating gradients through the network. Additionally, DGSM in our experiments used a smaller network than GANs, requiring roughly a quarter of sum and product operations to compute a single pass, without the need for any nonlinearities. This property makes DGSM specifically well suited for real-time robotics applications.

DC-GANs, being a convolutional model, lend themselves to very efficient implementations on GPUs. DGSM uses a more complicated network structure. However, in our current implementation in LibSPN, DGSM is real-time during inference and very efficient during learning, obtaining much faster inference than GANs. As a result, extending the model to include additional modalities and capture visual appearance or 3D structure is computationally feasible with DGSM.

The experiments with different inference types were all performed on the same model after a single training phase (separately for each dataset split). This demonstrates that our model spans not only multiple levels of abstraction, but also multiple tasks. In contrast, SVMs and GANs were optimized to solve a specific task. In particular, the model retains high capability to discriminate, outperforming a discriminative SVM. Yet, it is trained generatively to represent a joint distribution over low-level observations. Additionally, as demonstrated in the novelty detection experiments, it produces a useful, probabilistic uncertainty signal. Neither GANs nor SVMs explicitly represent likelihood of the data.

## VII. CONCLUSIONS AND FUTURE WORK

This paper presents DGSM, a unique generative spatial model, which to our knowledge, is the first application of Sum-Product Networks to the domain of robotics. Our results demonstrate that DGSM provides an efficient framework for learning deep probabilistic representations of robotic environments, spanning low-level features, geometry, and semantic representations. We have shown that DGSM has great generative and discriminative potential, and can be used to predict latent spatial concepts and missing observations. It can solve a variety of important robotic tasks, from semantic classification of places and uncertainty estimation, to novelty detection and generation of place appearances based on semantic information. DGSM has appealing properties and offers state-of-the-art performance. Our future efforts will focus on extending DGSM to include more complex perception provided by visual and depth sensors, as well as exploit the resulting deep representations for probabilistic reasoning and planning at multiple levels of abstraction.

## REFERENCES

[1] A. Aydemir, A. Pronobis, M. Göbelbecker, and P. Jensfelt, "Active visual object search in unknown environments using uncertain semantics," *T-RO*, vol. 29, no. 4, 2013.
[2] A. Pronobis and P. Jensfelt, "Large-scale semantic mapping and reasoning with heterogeneous modalities," in *ICRA*, 2012.
[3] Y. Bengio, A. Courville, and P. Vincent, "Representation learning: A review and new perspectives," *TPAMI*, vol. 35, no. 8, 2013.
[4] S. Levine, P. Pastor, A. Krizhevsky, and D. Quillen, "Learning hand-eye coordination for robotic grasping with deep learning and large-scale data collection." *ISER*, 2016.
[5] N. Sunderhauf, F. Dayoub, S. McMahon, B. Talbot, R. Schulz, P. Corke, G. Wyeth, B. Upcroft, and M. Milford, "Place categorization and semantic mapping on a mobile robot," in *ICRA*, 2016.
[6] R. Goeddel and E. Olson, "Learning semantic place labels from occupancy grids using CNNs," in *IROS*, 2016.
[7] H. Poon and P. Domingos, "Sum-product networks: A new deep architecture," in *UAI*, 2011.
[8] R. Peharz, R. Gens, F. Pernkopf, and P. Domingos, "On the latent variable interpretation in Sum-Product Networks," *TPAMI*, 2017.
[9] R. Gens and P. Domingos, "Discriminative learning of sum-product networks," in *NIPS*, 2012.
[10] R. Peharz, P. Robert, K. Georg, M. Pejman, and P. Franz, "Modeling speech with Sum-Product Networks: Application to bandwidth extension," in *ICASSP*, 2014.
[11] M. Amer and S. Todorovic, "Sum-Product Networks for activity recognition," *TPAMI*, vol. 38, no. 4, 2015.
[12] O. M. Mozos, C. Stachniss, and W. Burgard, "Supervised learning of places from range data using AdaBoost," in *ICRA*, 2005.
[13] A. Pronobis, O. M. Mozos, B. Caputo, and P. Jensfelt, "Multi-modal semantic place classification," *IJRR*, vol. 29, no. 2-3, Feb. 2010.
[14] I. Goodfellow, J. Pouget-Abadie, M. Mirza, B. Xu, D. Warde-Farley, S. Ozair, A. Courville, and Y. Bengio, "Generative adversarial nets," in *NIPS*, 2014.
[15] A. Radford, L. Metz, and S. Chintala, "Unsupervised representation learning with deep convolutional generative adversarial networks," arXiv:1511.06434 [cs.LG], 2015.
[16] A. Pronobis, A. Ranganath, and R. P. N. Rao, "LibSPN: A library for learning and inference with Sum-Product Networks and TensorFlow," in *ICML Workshop on Principled Approaches to Deep Learning*, 2017. [Online]. Available: http://www.libspn.org
[17] A. Oliva and A. Torralba, "Building the gist of a scene: The role of global image features in recognition," *Brain Research*, vol. 155, 2006.
[18] A. Ranganathan, "PLISS: Detecting and labeling places using online change-point detection," in *RSS*, 2010.
[19] P. Buschka and A. Saffiotti, "A virtual sensor for room detection," in *IROS*, 2002.
[20] A. Tapus and R. Siegwart, "Incremental robot mapping with fingerprints of places," in *ICRA*, 2005.
[21] Y. Sun, L. Bo, and D. Fox, "Attribute based object identification," in *ICRA*, 2013.
[22] A. Eitel, J. Springenberg, L. Spinello, M. Riedmiller, and W. Burgard, "Multimodal deep learning for robust RGB-D object recognition," in *IROS*, 2015.
[23] M. Madry, L. Bo, D. Kragic, and D. Fox, "ST-HMP: Unsupervised spatio-temporal feature learning for tactile data," in *ICRA*, 2014.
[24] D. Kingma and M. Welling, "Auto-encoding variational bayes," in *ICLR*, 2014.
[25] R. Yeh, C. Chen, T. Y. Lim, M. Hasegawa-Johnson, and M. N. Do, "Semantic image inpainting with perceptual and contextual losses," arXiv:1607.07539 [cs.CV], 2016.
[26] W.-C. Cheng, S. Kok, H. Pham, H. Chieu, and K. Chai, "Language modeling with Sum-Product Networks," in *Interspeech*, 2014.
[27] R. Gens and P. Domingos, "Learning the structure of Sum-Product Networks," in *ICML*, 2013.
[28] A. Rooshenas and D. Lowd, "Learning Sum-Product networks with direct and indirect variable interactions," in *ICML*, 2014.
[29] A. Darwiche, "A differential approach to inference in bayesian networks," *Journal of the ACM*, vol. 50, no. 3, May 2003.
[30] G. Grisetti, C. Stachniss, and W. Burgard, "Improved techniques for grid mapping with Rao-Blackwellized particle filters," *T-RO*, vol. 23, no. 1, 2007.